\documentclass{article}

\PassOptionsToPackage{numbers, compress}{natbib}
\usepackage[final]{nips_2017}

\usepackage[utf8]{inputenc} 
\usepackage[T1]{fontenc}    
\usepackage{hyperref}       
\usepackage{url}            
\usepackage{booktabs}       
\usepackage{amsfonts}       
\usepackage{nicefrac}       
\usepackage{microtype}      
\usepackage{ntheorem}       
\usepackage{cleveref}       
\usepackage{graphicx}       
\usepackage{subcaption}        
\usepackage[table,xcdraw]{xcolor} 

\graphicspath{{./figures/}}

\title{TequilaGAN: How to easily identify GAN samples}
\author{
  Rafael Valle \\
  CNMAT, UC Berkeley \\
  \texttt{rafaelvalle@berkeley.edu} \\
  \And
  Wilson Cai\\
  UC Berkeley\\
  \texttt{wcai@berkeley.edu} \\
  \And
  Anish Doshi\\
  UC Berkeley\\
  \texttt{apdoshi@berkeley.edu} \\
}

\begin{document}

\maketitle

\begin{abstract}
    In this paper we show strategies to easily identify fake samples generated
    with the Generative Adversarial Network framework. One strategy is based on
    the statistical analysis and comparison of raw pixel values and features
    extracted from them. The other strategy learns formal specifications from
    the real data and shows that fake samples violate the specifications of the
    real data. We show that fake samples produced with GANs have a universal
    signature that can be used to identify fake samples. We provide results on MNIST, CIFAR10, music and speech data.
\end{abstract}

\section{Introduction} \label{sec:introduction}
\textit{Fake samples} generated with the Generative Adversarial
Networks~\cite{goodfellow2014generative} framework have fooled humans and
machines to believe that they are indistinguishable from real samples.  Although
this might be true for the naked eye and the discriminator fooled by the
generator, it is unlikely that fake samples are numerically indistinguishable
from real samples. Inspired by formal methods, this paper focuses on the
evaluation of fake samples with respect to statistical summaries and formal
specifications computed on the real data. 

Since the Generative Adversarial Networks paper~\cite{goodfellow2014generative}
in 2014, most GAN related publications use a grid of image samples to accompany
theoretical and empirical results. Unlike Variational Autoencoders (VAEs) and
other models~\cite{goodfellow2014generative}, most of the evaluation of the
output of GAN trained Generators is qualitative: authors normally list higher
sample quality as one of the advantages of their method over other methods.
Although some numerical measures, like the inception score, are used to evaluate
GAN samples\cite{salimans2016improved}, interestingly, little is mentioned about
the numerical properties of fake samples and how these properties compare to
real samples.

In the context of Verified Artificial Intelligence\cite{seshia2016vai}, it is
hard to systematically verify that the output of a model satisfies the
specifications of the data it was trained on, specially when verification
depends on the existence of perceptually meaningful features. For example,
consider a model that generates images of humans: although it is possible to
compare color histograms of real and fake samples, we do not yet have robust
algorithms to verify if an image follows specifications derived from anatomy. 

This paper is related to the systematic verification of fake samples and focuses
on comparing numerical properties of fake and real samples. In addition to
comparing statistical summaries, we investigate how the Generator approximates
statistical modes in the real distribution and verify if the generated samples
violate specifications derived from the real distribution. We offer the
following main contributions: 
\begin{itemize}
\item We show that fake samples have properties that are barely noticed with
    visual of samples
\item We show that these properties can be used to identify the source of the
    data (real or fake)
\item We show that fake samples violate formal specifications learned from real
    data
\end{itemize}

\section{Related work}\label{sec:related_work}
Despite its youth, several publications (\cite{arjovsky2017towards},
\cite{salimans2016improved}, \cite{zhao2016energy},
\cite{radford2015unsupervised}) have investigated the use of the GAN framework
for sample generation and unsupervised feature learning.  Following the
procedure described in~\cite{breuleux2011quickly} and used
in~\cite{goodfellow2014generative}, earlier GAN papers evaluated the quality
of the fake samples by fitting a Gaussian Parzen window\footnote{Kernel
Density Estimation} to the fake samples and reporting the log-likelihood of
the test set under this distribution. It is known that this method has some
drawbacks, including its high variance and bad performance in high
dimensional spaces~\cite{goodfellow2014generative}. The inception score is
another widely adopted evaluation metric that fails to provide systematic
guidance on the evaluation of GAN models\cite{barratt2018note}.

Unlike other optimization problems, where analysis of the empirical risk is a
strong indicator of progress, in GANs the decrease in loss is not always
correlated with increase in image quality~\cite{arjovsky2017wasserstein}, and
thus authors still rely on visual inspection of generated images. Based on
visual inspection, authors confirm that they have not observed mode collapse or
that their framework is robust to mode collapse if some criteria is met
(\cite{arjovsky2017wasserstein}, \cite{gulrajani2017improved},
\cite{mao2016least}, \cite{radford2015unsupervised}).  In practice, github
issues where practitioners report mode collapse or not enough variety abound.

In their publications, \cite{mao2016least}, \cite{arjovsky2017wasserstein} and
\cite{gulrajani2017improved} propose alternative objective functions and
algorithms that circumvent problems that are common when using the original GAN
objective described in~\cite{goodfellow2014generative}. The problems addressed
include instability of learning, mode collapse and meaningful loss
curves~\cite{salimans2016improved}.

These alternatives do not eliminate the need or excitement\footnote{Despite of
authors promising on twitter to never train GANs again.} of visually inspecting
GAN samples during training, nor do they provide quantitative information about
the generated samples. 

\section{Methodology}\label{sec:method}
The experiments in this paper focus on three points: the first shows that fake
samples have properties that are hardly noticed with visual inspection and that
are tightly related to the requirements of differentiability; the second shows
that there are numerical differences between statistical moments computed on
features extracted from real and fake samples that can be used to identify the
data; the third shows that fake samples violate formal specifications learned
from the real data. In the following subsections we enumerate the datasets,
features and GAN frameworks herein used.

\subsection{Datasets}
In our experiments, we use MNIST, CIFAR10, a MIDI dataset of 389 Bach Chorales
downloaded from the web and a subsample of the NIST 2004 telephone
conversational speech dataset\cite{cai2018attacking}.

\subsection{Features}
\begin{figure}[!h]
    \centering
    \begin{subfigure}[b]{0.4\textwidth}
        \includegraphics[width=\linewidth]{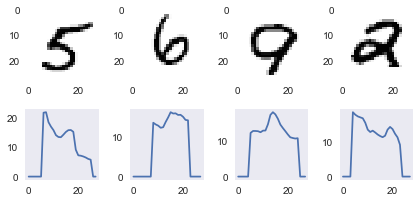}
        \caption{MNIST samples and centroids}
        \label{fig:mnist_centroids}
    \end{subfigure}
    \quad
    \begin{subfigure}[b]{0.4\textwidth}
        \includegraphics[width=\linewidth]{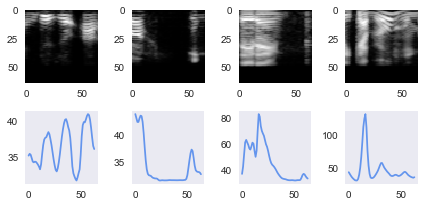}
        \caption{Mel-Spectrograms and centroids}
        \label{fig:spectrogram_centroids}
    \end{subfigure}
    \caption{Spectral centroids on digits and Mel-Spectrograms}
    \label{fig:centroids}
\end{figure}

The \textbf{spectral centroid}~\cite{peeters2004large} is a feature commonly used in the
audio domain, where it represents the barycenter of the spectrum. This feature
can be applied to other domains and we invite the reader to visualize 
Figure~\ref{fig:centroids} for examples on MNIST and
Mel-Spectrograms~\cite{peeters2004large}. For each column in an image, we 
transform the pixel values into row probabilities by normalizing them by the
column sum, after which we take the expected row value, thus obtaining the
spectral centroid.

\begin{figure}[!h]
    \centering
    \begin{subfigure}[b]{0.4\textwidth}
        \includegraphics[width=\linewidth]{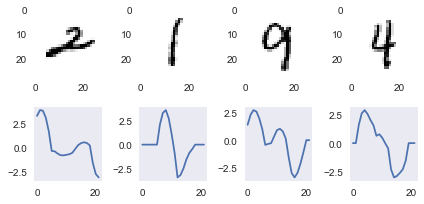}
        \caption{MNIST samples and slopes}
        \label{fig:mnist_slopes}
    \end{subfigure}
    \quad
    \begin{subfigure}[b]{0.4\textwidth}
        \includegraphics[width=\linewidth]{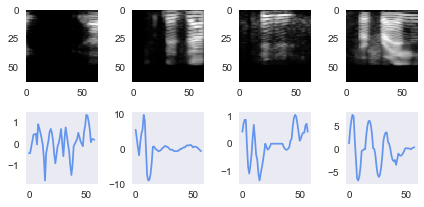}
        \caption{Mel-Spectrograms and slopes}
        \label{fig:spectrogram_slopes}
    \end{subfigure}
    \caption{Spectral slopes on digits and Mel-Spectrograms}
    \label{fig:slopes}
\end{figure}

    The \textbf{spectral slope} adapted from~\cite{peeters2004large} is computed by applying linear regression using an overlapping
sliding window of size 7. For each window, we regress the spectral centroids on
the column number \textit{mod} the window size. Figure~\ref{fig:slopes} shows these
features computed on MNIST and Mel-Spectrograms.

\subsection{GAN Frameworks}
We investigate samples produced with the DCGAN architecture using the
Least-Squares GAN (LSGAN)~\cite{mao2016least} and the improved Wasserstein GAN
(IWGAN/WGAN-GP)~\cite{gulrajani2017improved}. We also compare adversarial MNIST samples
produced with the fast gradient sign method
(FGSM)~\cite{goodfellow2014explaining}. We evaluate the normally used
non-linearities, sigmoid and tanh, on the output of the generator \textbf{and}
and other transformations such as the scaled tanh and identity.

\section{Experiments}\label{sec:experiments}
\subsection{MNIST}
The experiment focuses on showing numerical properties of fake MNIST samples and
features therein, unknown to the naked eye, that can be used to identify them as
produced by a GAN. 

We start by comparing the distribution of features computed over the MNIST
training set to other datasets, including the MNIST test set, samples generated
with GANs and adversarial samples computed using the FGSM. The training data is
scaled to $[0, 1]$ and the random baseline is sampled from a Bernoulli
distribution with probability equal to the mean value of pixel intensities in
the MNIST training data, 0.13. Each GAN model is trained until the loss plateaus
and the generated samples look similar to the real samples. The datasets
compared have 10 thousand samples each.

\begin{figure}[!h]
  \begin{center}
  \includegraphics[width=0.8\linewidth]{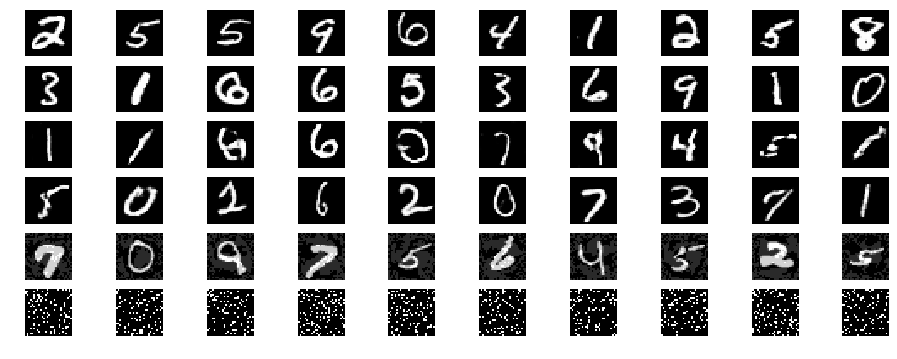}
  \caption{Samples drawn from MNIST train, test,
LSGAN, IWGAN, FSGM and Bernoulli}
  \label{fig:mnist_samples}
  \end{center}
\end{figure}

Visual inspection of the generated samples in
Figure~\ref{fig:mnist_samples} show that IWGAN seems to produce better samples
than LSGAN. Quantitatively, we use the MNIST training set as a reference and compare the
distribution of pixel intensities.  Table~\ref{tbl:mnist_pixel} reveals that
although samples generated with LSGAN and IWGAN look similar to the training
set, they are considerably different from the training set given the Kolgomorov-Smirnov (KS) Two
Sample Test and the Jensen-Shannon Divergence (JSD), specially with respect to
the same statistics on the MNIST test data. 

\begin{table}[!h]
\centering
\begin{tabular}{l|ll|l|}
                   & \multicolumn{2}{c|}{\cellcolor[HTML]{C0C0C0}KS Two Sample Test} & \multicolumn{1}{c|}{\cellcolor[HTML]{C0C0C0}JSD} \\
                   & Statistic   & P-Value   &                \\
mnist\_train       & 0.0         & 1.0       & 0.0            \\
mnist\_test        & 0.003177    & 0.0       & 0.000029       \\
mnist\_lsgan       & 0.808119    & 0.0       & 0.013517       \\
mnist\_iwgan       & 0.701573    & 0.0       & 0.014662       \\
mnist\_adversarial & 0.419338    & 0.0       & 0.581769       \\
mnist\_bernoulli   & 0.130855    & 0.0       & 0.0785009      
\end{tabular}
\caption{Statistical comparison over the distribution of pixel values for
different samples using MNIST training set as reference.}
\label{tbl:mnist_pixel}
\end{table}

\begin{figure}[!h]
  \includegraphics[width=\linewidth]{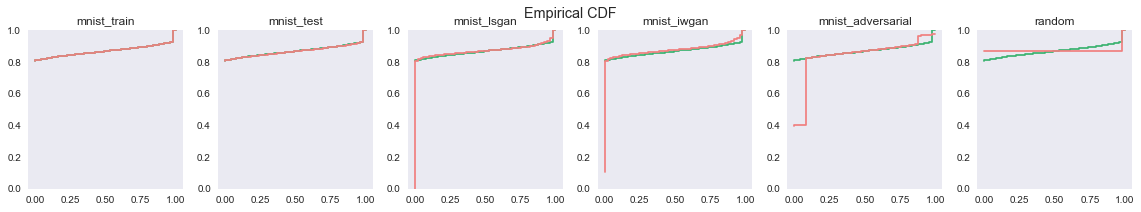}
  \caption{Pixel empirical CDF of training data as reference (green) and other datasets(red)}
  \label{fig:mnist_pixel_ecdf}
\end{figure}

These numerical phenomena can be understood by investigating the empirical CDFs
in Figure~\ref{fig:mnist_pixel_ecdf}. The distribution of pixel values of the
samples generated with the GAN framework is mainly bi-modal and asymptotically
approaches the modes of the distribution in the real data, values $0$ and $1$.
Expectedly, the FGSM method, noted as \textit{mnist\_adversarial}, causes a shift on the
modes of the distribution that can be easily identified.

\begin{figure}[!h]
  \includegraphics[width=\linewidth]{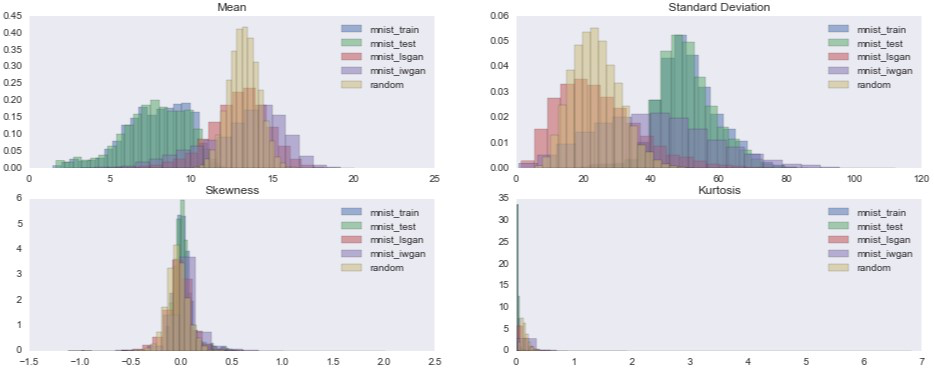}
  \caption{Distribution of moments of spectral centroids computed on each image.}
  \label{fig:mnist_centroid_distribution}
\end{figure}

In addition, plots of the distribution of statistical moments of the spectral
centroid in ~\ref{fig:mnist_centroid_distribution} suggests that the fake images
are more noisy than the real images. Consider for example images produced by
randomly sampling a Bernoulli distribution with parameter estimated from the
training data. These images have pixel values of $0$ or $1$ that are equally
distributed \footnote{This spatial distribution is independent of the parameter
of the Bernoulli distribution.} over the image. Well, an image that has pixels
values distributed in such a manner will have a distribution of mean spectral
centroid with a mode at the center row of the image. This and the similarity
between the distribution of mean spectral centroids from fake data and Bernoulli
data suggest that the fake images have noise that are also equally spatially
distributed.

\begin{figure}[!h]
  \includegraphics[width=\linewidth]{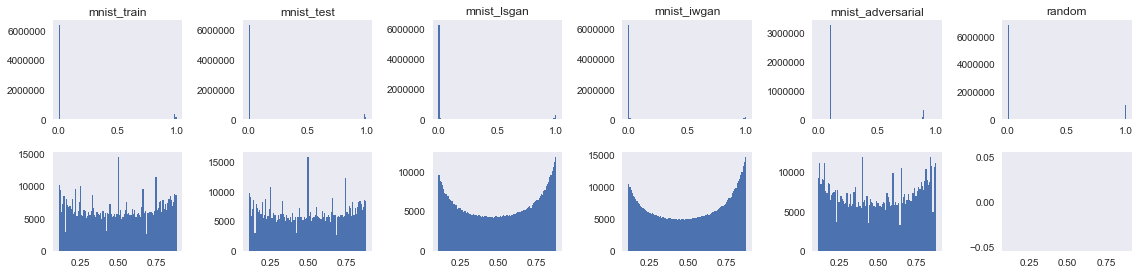}
  \caption{Histogram of pixel values for each dataset. First row shows
    values within [0, 1] and 100 bins. Second row shows
    values within [0.11, and 0.88] and 100 bins.}
  \label{fig:mnist_pixel_distribution}
\end{figure}

Last, Figure~\ref{fig:mnist_pixel_distribution} shows that the GAN generated
samples smoothly approximate the modes of the distribution. This smooth
approximation is considerably different from the training and test sets.
Although not perceptually meaningful, these properties can be used to identify
the source of the data.

Our first hypothesis for the smooth approximation of the modes of the
distribution was that it would be present in any data produced with a generator
that is trained using stochastic gradient descent \textbf{and} an asymptotically
converging activation function, such as sigmoid or tanh, at the output of the
generator. To evaluate this hypothesis, we conducted a set of experiments using
different GAN architectures (WGAN-GP, LSGAN, DCGAN) with different activation
functions, including linear and the scaled tanh, at the output of the Generator,
keeping the Discriminator fixed.

\begin{figure}[!h]
  \centering
  \begin{subfigure}[b]{0.49\textwidth}
    \includegraphics[width=0.49\linewidth]{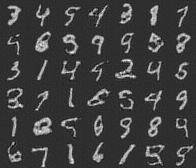}
    \includegraphics[width=0.49\linewidth]{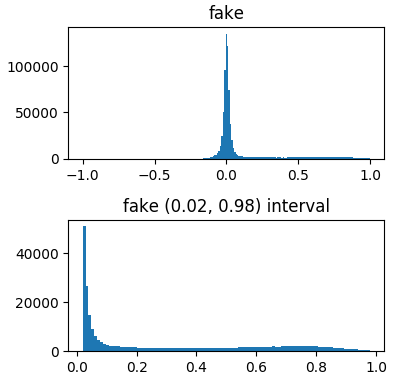}
    \caption{Linear}
  \end{subfigure}
  \begin{subfigure}[b]{0.49\textwidth}
    \includegraphics[width=0.49\linewidth]{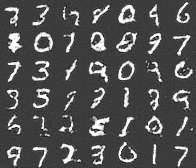}
    \includegraphics[width=0.49\linewidth]{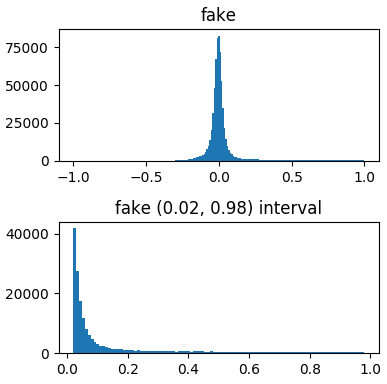}
    \caption{Scaled Tanh}
  \end{subfigure}
  \caption{Fake MNIST samples and pixel distribution from generators trained with DCGAN, Batch Norm and linear or scaled tanh activation functions.}
  \label{fig:mnist_gan_activations}
\end{figure}

To our surprise, we noticed that the models trained with linear or scaled tanh
activations were partially able to produce images that were similar to the MNIST
training data and the distribution of pixel intensities, although uni-modal
around zero, still possessed a smooth looking curve. This is illustrated in
Figure~\ref{fig:mnist_gan_activations}.

\begin{figure}[!h]
  \centering  
  \begin{subfigure}[b]{0.49\textwidth}
    \includegraphics[width=0.49\linewidth]{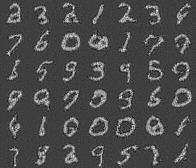}
    \includegraphics[width=0.49\linewidth]{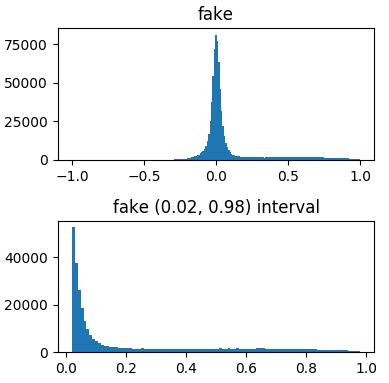}
    \caption{DCGAN Linear, Binary data}        
  \end{subfigure}
  \begin{subfigure}[b]{0.49\textwidth}
    \includegraphics[width=0.49\linewidth]{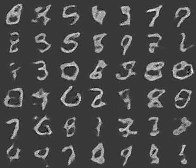}
    \includegraphics[width=0.49\linewidth]{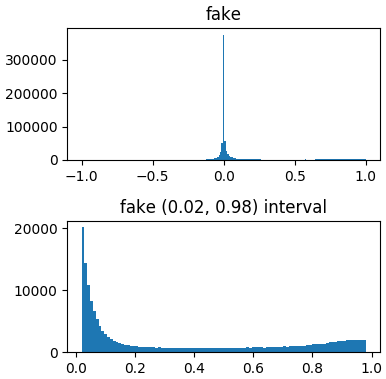}
    \caption{WGAN-GP Linear, Binary data}          
  \end{subfigure}
  \caption{Fake MNIST samples and pixel distribution from generators trained on binarized real data with DCGAN and WGAN-GP, Batch Norm and linear activation functions.}
  \label{fig:mnist_gan_binary_data}
\end{figure}

We then postulated that the smooth behavior was due to smoothness in the pixels
intensities of the training data itself. To validate this, we binarized the real
data by first scaling it between $[0, 1]$ and then thresholding it at $0.5$.
With this alteration the distribution of the pixel intensities of the real data
becomes completely bi-modal with modes at $0$ and $1$.
Figure~\ref{fig:mnist_gan_binary_data} shows that, as expected, the smooth
behavior remained. 

With this empirical evidence at hand, we provide an informal analysis of the
smoothness of the distribution of pixels of the generated data from the
perspective of optimization, differentiation and function approximation with
neural networks. We know that backpropagation and stochastic gradient descent
are used to update the weights of a neural network model based on the gradient
of the loss with respect to the weights. We also know that differentiation
requires the function that is being differentiated to have some high degree of
smoothness and be differentiable almost everywhere. Hence, we conjecture that
the inductive bias of this learning setup is that of smoothness given the
requirements of differentiation.
 
We also speculate that this inductive bias is responsible for the smoothness of
the distribution of pixel values at any iteration during training and that the U
shape of the distribution of pixel values, similar to blurring, is the byproduct
of an smooth approximation of the function that is being
learned\footnote{Consider approximating a function with polynomials of
increasing degrees}.

\subsection{CIFAR10}
Expecting that the results we obtained during our MNIST experiments would
generalize to other images, we briefly investigate the properties of CIFAR10
fake samples generated with the IWGAN framework and using the DCGAN
architecture. The models were trained using the CIFAR10's training set with 50k
samples and following the
\href{https://github.com/igul222/improved_wgan_training/blob/master/gan_cifar.py}{experimental
setup} provided by the IWGAN authors.

\begin{figure}[!h]
    \begin{subfigure}[b]{0.65\textwidth}
        \includegraphics[width=0.8\linewidth]{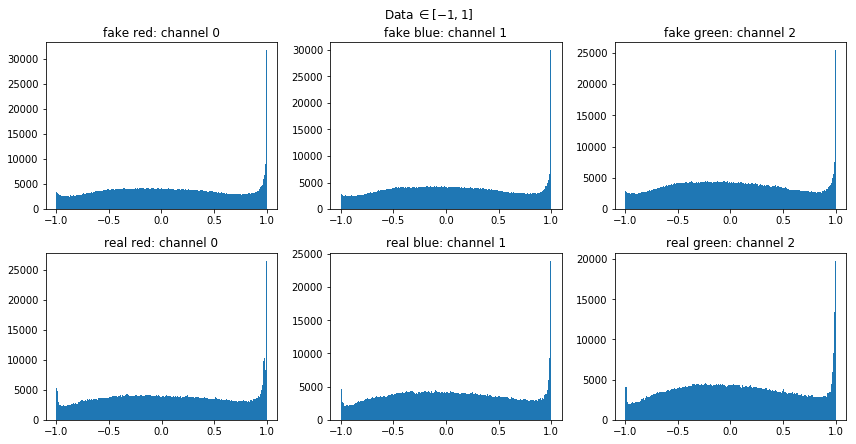}
        \caption{Distribution of pixel intensities from real and fake samples}
        \label{fig:cifar10_dist}
    \end{subfigure}
    \quad
    \begin{subfigure}[b]{0.3\textwidth}
        \resizebox{1.0\textwidth}{!}{%
        \begin{tabular}{l|ll|l|}
                           & \multicolumn{2}{c|}{\cellcolor[HTML]{C0C0C0}KS Two Sample Test} & \multicolumn{1}{c|}{\cellcolor[HTML]{C0C0C0}JSD} \\
                           & Statistic   & P-Value  &         \\
        train       & 0.0         & 1.0   & 0.0       \\
        test        & 0.0061      & 0.0   & 0.00005   \\
        iwgan       & 0.0107      & 0.0   & 0.21378   \\
        \end{tabular}
        }
        \caption{Test statistics}
        \label{tbl:cifar10}
    \end{subfigure}
\end{figure}

We report results on CIFAR10 train, test and IWGAN generated samples, all with
ten thousand items each. An informal analysis of Figure~\ref{fig:cifar10_dist}
shows that the distribution of pixels per channel is different between the real
and fake data, specially for pixels values close to $-1$. Numerically, we can
see in Table~\ref{tbl:cifar10} that the JSD between the samples from the
training data and IWGAN samples is considerably large with respect to the same
statistics on the test data. 

\begin{figure}[!h]
  \begin{center}
  \includegraphics[width=0.6\linewidth]{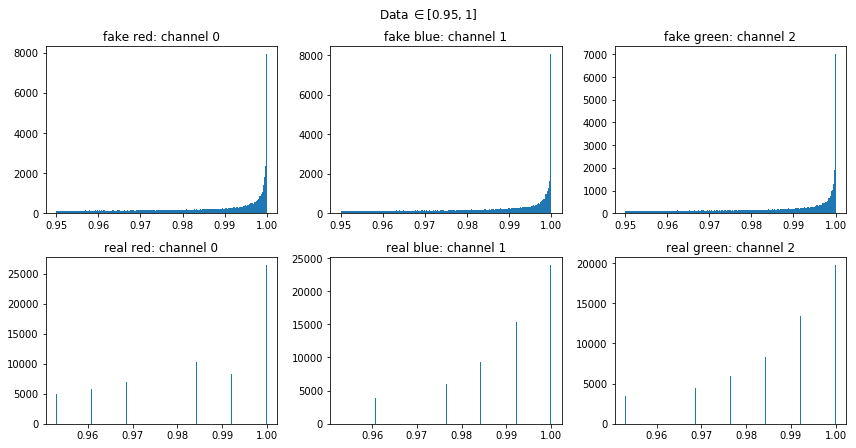}
  \caption{Distribution of pixel intensities from real and fake samples. The
      pixel distribution computed over fake samples is continuous and smoothly
      approximates the mode at 1.0}
  \label{fig:cifar10_dist_right}
  \end{center}
\end{figure}

Figure~\ref{fig:cifar10_dist_right} shows a behavior seen in our MNIST
experiments: the GAN generated samples smoothly approximate the mode of the
pixel value distribution at $1$ and this smooth approximation is considerably
different from the training set. As we previously explained, these properties
can be used to identify the gan samples although they might be not perceptible
with the naked eye.

\subsection{Bach Chorales}
We investigate the properties of Bach chorales generated with the GAN framework
and verify if they satisfy musical specifications learned from real data.  Bach
chorales are polyphonic pieces of music, normally written for 4 or 5 voices,
that follow a set of specifications or rules\footnote{The specifications define
the characteristics of the musical style.}. For example, a global specification
could assert that only a set of durations are valid; a local specification could
assert that only certain transitions between states (notes) are valid depending
on the current harmony.

For this experiment, we convert the dataset of Bach chorales to piano rolls. The
piano roll is a representation in which the rows represent note numbers, the
columns represent time steps and the cell values represent note intensities. We
compare the distribution of features computed over the training set, test set,
GAN generated samples and a random baseline sampled from a Bernoulli
distribution with probability equal to the normalized mean value of intensities
in the training data. After scaling and thresholding, the intensities in the
training and test data are strictly bi-modal and equal to $0$ or $1$.
Figure~\ref{fig:chorales_samples} below shows training, test, IWGAN and
Bernoulli samples, with modes on 0 and 1. Each dataset has approximately 1000
image patches.

\begin{figure}[!h]
  \begin{center}
  \includegraphics[width=0.8\linewidth]{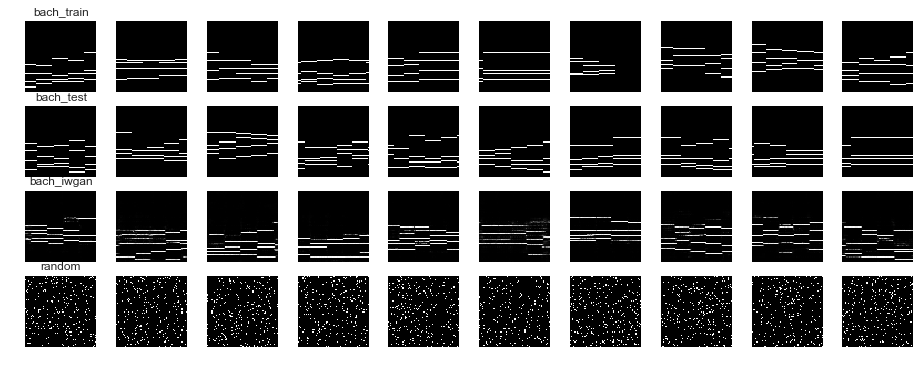}
  \caption{Samples drawn from Bach Chorales train, test,
IWGAN, and Bernoulli respectively.}
  \label{fig:chorales_samples}
  \end{center}
\end{figure}

\begin{figure}[!h]
  \includegraphics[width=\linewidth]{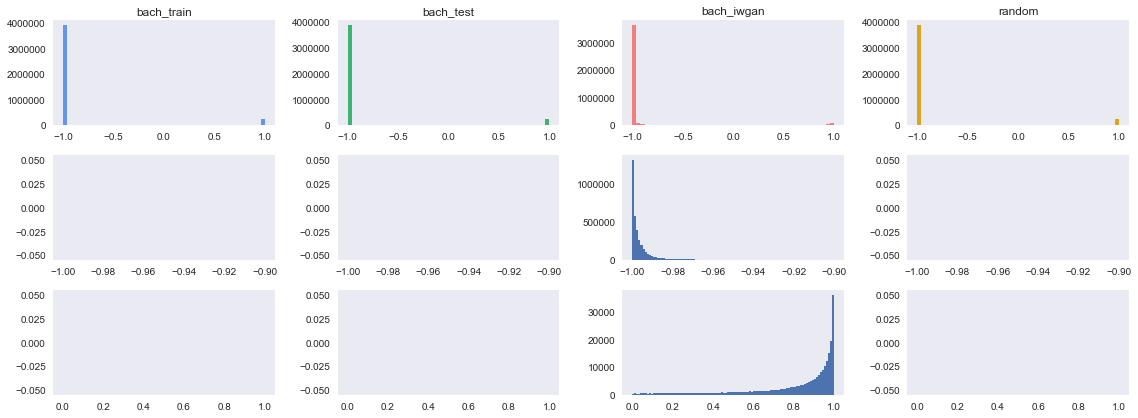}
  \caption{}
  \label{fig:chorales_intensity_distribution}
\end{figure}

Figure~\ref{fig:chorales_intensity_distribution} shows a behavior that is
similar to our previous MNIST experiments: the IWGAN asymptotically approximates
the modes of the distribution of intensity values. 

Following, we investigate if the generated samples violate the specifications of
Bach chorales. We use these piano rolls to compute boolean
Chroma~\cite{peeters2004large} features and to compute an empirical Chroma
transition matrix, where the positive entries represent existing and valid
transitions. The transition matrix built on the training data is taken as the
reference specification, i.e. anything that is not included is a violation of
the specification. Table~\ref{tbl:chroma_violations} shows the number of
violations given each dataset. 

Although Figure~\ref{fig:chorales_samples} shows generated samples that look
similar to the real data, the IWGAN samples have over 5000 violations, 10 times
more than the test set! Violation of specifications is a strong evidence that
fake samples do not come from the same distribution as real data. Furthermore,
to the trained ear the fake samples violate the style of Bach. We invite readers
to listen to
\href{https://soundcloud.com/d_alma/sets/improved-wasserstein-gans-bach}{
them}.

\begin{table}[!h]
\centering
\begin{tabular}{lllll}
& \cellcolor[HTML]{C0C0C0}bach\_train & \cellcolor[HTML]{C0C0C0}bach\_test & \cellcolor[HTML]{C0C0C0}bach\_iwgan & \cellcolor[HTML]{C0C0C0}bach\_bernoulli \\ \cline{2-5} 
\multicolumn{1}{l|}{\cellcolor[HTML]{C0C0C0}Number of Violations} & 0                                   & 429                                & 5029                                & 58284                                  
\end{tabular}
\caption{Number of specification violations with training data as
    reference.}
\label{tbl:chroma_violations}
\end{table}

\begin{figure}[!h]
    \begin{subfigure}[b]{0.65\textwidth}
        \includegraphics[width=\linewidth]{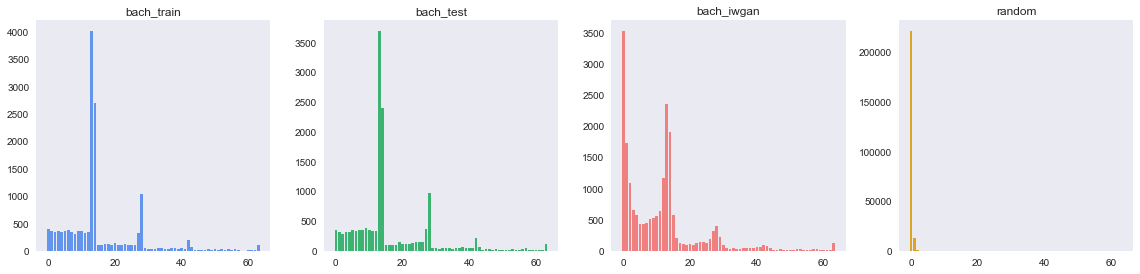}
        \caption{Histogram of note durations}
        \label{fig:chorales_duration_distribution}
    \end{subfigure}
    \quad
    \begin{subfigure}[b]{0.3\textwidth}
        \resizebox{1.0\textwidth}{!}{%
        \begin{tabular}{l|ll|l|}
                           & \multicolumn{2}{c|}{\cellcolor[HTML]{C0C0C0}KS Two Sample Test} & \multicolumn{1}{c|}{\cellcolor[HTML]{C0C0C0}JSD} \\
                           & Statistic   & P-Value  &         \\
        train       & 0.0          & 1.0      & 0.0     \\
        test        & 0.09375      & 0.929    & 0.002   \\
        iwgan       & 0.21875      & 0.080    & 0.084   \\
        bernoulli   & 0.93750      & 0.0      & 0.604   
        \end{tabular}
        }
        \caption{Test statistics}
        \label{tbl:duration}
    \end{subfigure}
\end{figure}

In addition to experiments with Chroma features, we computed the distribution of
note durations on the boolean piano roll described above.
Figure~\ref{fig:chorales_duration_distribution} shows the distribution of
note durations within each dataset. The train and test data are approximately
bi-modal and, again, the improved WGAN smoothly approximates the
dominating modes of the distribution. Table~\ref{tbl:duration} provides a numerical 
comparison between datasets.

\subsection{Speech}
\begin{figure}[!h]
  \begin{center}
  \includegraphics[width=0.8\linewidth]{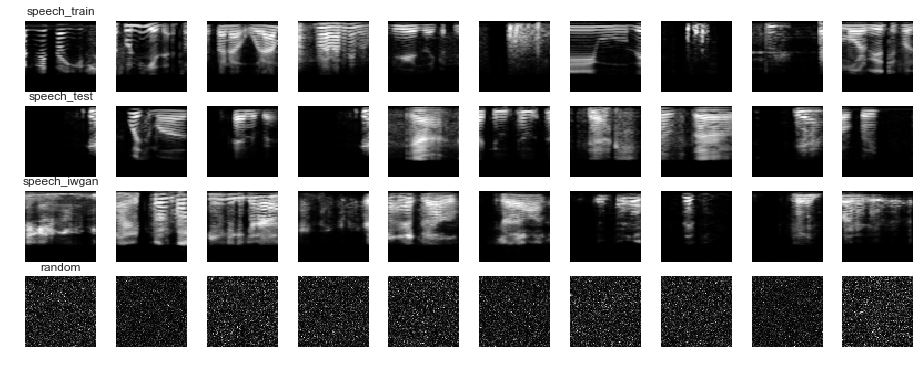}
  \caption{Samples drawn from NIS2004 train, test, IWGAN, and exponential
      respectively.}
  \label{fig:speech_samples}
  \end{center}
\end{figure}

Within the speech domain, we investigate Mel-Spectrogram real and fake samples.
We divide the NIST 2004 dataset into training and test set, generate samples
with the GAN framework and use a random baseline sampled from a Exponential
distribution with parameters chosen using heuristics.  The generated samples can
be seen in Figure~\ref{fig:speech_samples}.  We obtain the Mel-Spectrogram by
projecting a spectrogram onto a mel scale, which we do with the python library
librosa~\cite{mcfee2015librosa}. More specifically,  we project the spectrogram
onto 64 mel bands, with window size equal to 1024 samples and hop size equal to
160 samples, i.e. frames of 100ms long. Dynamic range compression is computed
as described in~\cite{lukic2016speaker}, with $log(1 + C*M)$, where $C$ is the
compression constant scalar set to $1000$ and $M$ is the matrix representing the
Mel-Spectrogram.  Each dataset has approximately 1000 image patches and the GAN
models are trained using DCGAN with the improved Wasserstein GAN algorithm.

\begin{figure}[!h]
    \begin{subfigure}[b]{0.65\textwidth}
        \includegraphics[width=1.0\linewidth]{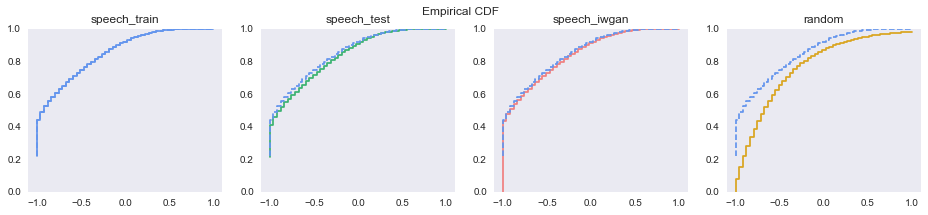}
        \caption{Empirical CDF computed on intensities. Training data in blue
        and other datasets.}
        \label{fig:speech_intensity_ecdf}
    \end{subfigure}
    \quad
    \begin{subfigure}[b]{0.3\textwidth}
        \resizebox{1.0\textwidth}{!}{%
        \begin{tabular}{l|ll|l|}
                           & \multicolumn{2}{c|}{\cellcolor[HTML]{C0C0C0}KS Two Sample Test} & \multicolumn{1}{c|}{\cellcolor[HTML]{C0C0C0}JSD} \\
                    & Statistic    & P-Value  &         \\
        train       & 0.0          & 1.0      & 0.0     \\
        test        & 0.03685      & 0.0      & 0.00080   \\
        iwgan       & 0.22149      & 0.0      & 0.00056   \\
        bernoulli   & 0.36205      & 0.0      & 0.11423   
        \end{tabular}
        }
        \caption{Test statistics}
        \label{tbl:speech_intensity}
    \end{subfigure}
    \caption{Empirical CDF and statistical tests of speech intensity}
    \label{fig:speech_intensity}
\end{figure}

Figure~\ref{fig:speech_intensity_ecdf} shows the empirical CDFs of intensity
values. Unlike our previous experiments where intensities (Bach Chorales) or
pixel values (MNIST, CIFAR10) were linear and discrete, in this experiment
intensities are continuous and compressed using the log function.  This
considerably reduces the distance between the empirical CDFs of the training
data and GAN samples, specially around the saturating points of the tanh
non-linearity, $-1$ and $1$ in this case.

Table~\ref{tbl:speech_intensity} shows a significant difference between the
KS-Statistic of test samples and fake samples with respect to the training data.
However, an adversary can manipulate the fake samples to considerably decrease
this difference and still keep the high similarity in features harder to
simulate such as moments of spectral centroid or slope. 

\begin{figure}[!h]
    \centering
    \begin{subfigure}[b]{0.45\textwidth}
        \includegraphics[width=\linewidth]{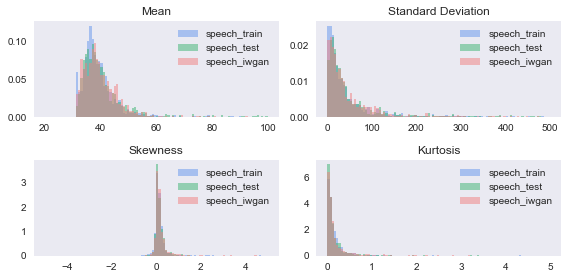}
        \caption{Spectral Centroid Moments}
        \label{fig:speech_spectral_centroid_moments}
    \end{subfigure}
    \quad
    \begin{subfigure}[b]{0.45\textwidth}
        \includegraphics[width=\linewidth]{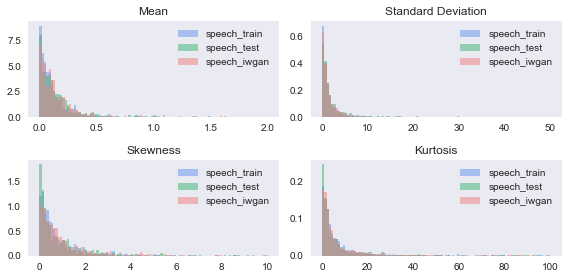}
        \caption{Spectral Slope Moments}
        \label{fig:speech_spectral_slope_moments}
    \end{subfigure}
    \caption{Moments of spectral centroid (left) and slope(right)}
    \label{fig:speech_moments}
\end{figure}

Figure~\ref{fig:speech_moments} shows the distribution of statistical moments computed on
spectral centroids and slope. The distributions from different sources
considerably overlap, indicating that the generator has efficiently approximated
the real distribution of these features.

\section{Conclusions}\label{sec:conclusions}
In this paper we investigated numerical properties of samples produced with
adversarial methods, specially Generative Adversarial Networks. We showed that
fake samples have properties that are barely noticed with visual of samples,
namely the fact that, due to stochastic gradient descent and the requirements of
differentiability, fake samples smoothly approximate the dominating modes of the
distribution. We analysed statistical measures of divergence between real data
and other data and the results showed that even in simple cases, e.g.
distribution of pixel intensities, the divergence between training data and fake
data is large with respect to test data. Finally, we mined specifications from
real data and showed that, unlike test data, the fake data considerably violates
the specifications of the real data.

In the context of adversarial attacks, these large differences in distribution
and specially violations of specification can be used to identify data that is
fake. In our results we show that, although some of the features used to learn
specifications in this paper are weakly perceptually correlated with the content
of the image, they certainly can be used to identify fake samples.

Although not common practice, one could possibly circumvent the difference in
support between the real and fake data by training Generators that explicitly
sample a distribution that replicates the support of the real data, i.e. 256
values in the case of discretized images. Conversely, one could mine
specifications that are easy to learn from real data but hardly differentiable.
These are topics that are not limited to GANs and remain to be explored in the
larger domain of Verified Artificial Intelligence\cite{seshia2016vai}.

\subsubsection*{Acknowledgments}
We are thankful to Ryan Prenger and Kevin Shih for their feedback on this paper.
We acknowledge NVIDIA for providing us with the Titan X GPU used in these
experiments.

\bibliographystyle{plain}
\bibliography{references}
\end{document}